\documentclass[10pt,conference]{IEEEtran}  % Comment this line out
\usepackage{amsmath,amssymb,amsfonts}
\usepackage{graphicx}
\usepackage{textcomp}
\usepackage[table,xcdraw]{xcolor}
\usepackage{booktabs}
\usepackage{diagbox}
\usepackage{algorithm}
\usepackage{algorithmicx}
\usepackage{algpseudocode}
\usepackage[table,xcdraw]{xcolor}
\usepackage{rotating}

\usepackage{colortbl}
\usepackage{multirow}
\usepackage{caption}
\usepackage{stfloats}
\usepackage{geometry}
\usepackage{makecell}

\usepackage{titlesec}
\usepackage[numbers,square,sort]{natbib}

\def\BibTeX{{\rm B\kern-.05em{\sc i\kern-.025em b}\kern-.08em
    T\kern-.1667em\lower.7ex\hbox{E}\kern-.125emX}}

\begin{document}

% \newgeometry{top=72pt, left=54pt, right=54pt, bottom=54pt}

\title{Task-Aware Parameter-Efficient Fine-Tuning of Large Pre-Trained Models at the Edge}
% \vspace{-6mm}

\author{
    \IEEEauthorblockN{Senkang Hu${}^{\star,\dagger}$, Yanan Ma$^{\star,\dagger }$, Yihang Tao${}^{\star,\dagger}$, Zhengru Fang$^{\star,\dagger}$, Zihan Fang${}^{\star,\dagger}$, Yiqin Deng$^{\star,\dagger}$, \\Sam Kwong$^\ddagger $, Yuguang Fang${}^{\star,\dagger}$\\
    $^\star$Hong Kong JC STEM Lab of Smart City, Hong Kong SAR.\\
    $^\dagger$Department of Computer Science, City University of Hong Kong, Hong Kong SAR.\\
    $^\ddagger$Department of Computing and Decision Sciences, Lingnan University, Hong Kong SAR.\\
    Email: \{senkang.forest, yananma8-c, yihang.tommy, zhefang4-c, zihanfang3-c\}@my.cityu.edu.hk, \\
    samkwong@ln.edu.hk, \{yiqideng, my.Fang\}@cityu.edu.hk\\
    }
    \vspace{-10mm}
    }
% \vspace*{72pt}
\maketitle

\newcommand\blfootnote[1]{%
\begingroup
\renewcommand\thefootnote{}\footnote{#1}%
\addtocounter{footnote}{-1}%
\endgroup
}

% \vspace{-3.7mm}
\begin{abstract}
Large language models (LLMs) have achieved remarkable success in various tasks, such as decision-making, reasoning, and question answering. They have been widely used in edge devices. However, fine-tuning LLMs to specific tasks at the edge is challenging due to the high computational cost and the limited storage and energy resources at the edge. To address this issue, we propose TaskEdge, a task-aware parameter-efficient fine-tuning framework at the edge, which allocates the most effective parameters to the target task and only updates the task-specific parameters. Specifically, we first design a parameter importance calculation criterion that incorporates both weights and input activations into the computation of weight importance. Then, we propose a model-agnostic task-specific parameter allocation algorithm to ensure that task-specific parameters are distributed evenly across the model, rather than being concentrated in specific regions. In doing so, TaskEdge can significantly reduce the computational cost and memory usage while maintaining performance on the target downstream tasks by updating less than 0.1\% of the parameters. In addition, TaskEdge can be easily integrated with structured sparsity to enable acceleration by NVIDIA's specialized sparse tensor cores, and it can be seamlessly integrated with LoRA to enable efficient sparse low-rank adaptation. Extensive experiments on various tasks demonstrate the effectiveness of TaskEdge.
\end{abstract}

\section{Introduction}

Large pre-trained models, represented by large language models (LLMs), have emerged as pioneering use cases of artificial general intelligence (AGI), demonstrating remarkable capabilities across diverse domains including question answering, complex reasoning, content summarization, code generation, and creative writing. These models have revolutionized natural language processing by exhibiting unprecedented understanding of context, nuance, and human intent. For example, DeepSeek-R1 \cite{deepseekai2025deepseekr1incentivizingreasoningcapability} demonstrates powerful reasoning capabilities and the ability to solve complex problems.
Furthermore, multimodal large models have achieved significant breakthroughs in image understanding, video comprehension, speech recognition, and speech generation, enabling seamless integration of different data modalities and expanding the frontier of AI applications. For example, OpenAI o1 \cite{zhong2024evaluationopenaio1opportunities} is a multimodal model that accepts image and text inputs and produces text outputs exhibiting human-level performance on various benchmarks. Apart from these general-purpose models, there are many task-specific models that are fine-tuned on specific downstream tasks. For example, LMDrive \cite{Shao_2024_CVPR} is a fine-tuned multimodal model specifically designed for autonomous driving. AnomalyGPT \cite{guAnomalyGPTDetectingIndustrial2024} is fine-tuned for industrial anomaly detection.
However, large pre-trained models typically contain billions of parameters, making them extremely resource-intensive and computationally demanding. For instance, pre-training a LLaMA-7B model from scratch with just a single batch size consumes at least 58 GB of memory: 14GB for the trainable parameters themselves, 42GB for Adam optimizer states and weight gradients, and an additional 2GB for activations \cite{10.5555/3692070.3694598}. 
Such memory demands render training infeasible on consumer-grade hardware like the NVIDIA RTX 4090, which offers only 24GB of memory, creating significant barriers for fine-tuning on edge devices. This capability is crucial for enabling privacy, as sensitive data, like health records or personal habits, stays on the device, reducing breach risks during transmission. It is also vital to enable real-time adaptation, allowing the model to learn from new data instantly in applications such as autonomous driving.

To tackle these resource constraints, parameter-efficient fine-tuning (PEFT) methods have emerged as a promising solution, enabling adaptation of large pre-trained models to specific downstream tasks while updating only a small subset of parameters \cite{guo-etal-2021-parameter}. 
A notable example is Low-Rank Adaptation (LoRA) \cite{hu2022lora}, which introduces trainable low-rank matrices to modify weights without changing the original architecture. This approach significantly reduces memory and computational costs while maintaining reasonable performance, making it suitable for resource-constrained edge computing environments.

Despite its efficiency benefits, LoRA faces important limitations. Recent studies show that LoRA often cannot match the performance of full fine-tuning \cite{xia2024chainloraefficientfinetuning}, particularly for complex tasks. Additionally, when applied to pre-training scenarios, researchers have found that LoRA requires an initial full-rank model training phase before transitioning to low-rank optimization \cite{lialin2024relora}. These limitations likely stem from two fundamental issues: first, the optimal weight updates for many tasks may not naturally fit within a low-rank structure; second, the low-rank reparameterization fundamentally alters gradient flow during training, potentially limiting the model's ability to learn certain patterns.

In order to address these challenges, we propose TaskEdge, a novel task-aware parameter-efficient fine-tuning framework at the edge, which can allocate the most effective parameters to the target task and only update the task-specific parameters, thereby significantly reducing the computational cost, while maintaining the performance on the target downstream task. 

Specifically, we first design a task-specific parameter importance metric that combines both weight values and input activations. For each weight $(i,j)$ in layer $k$, we compute an importance score by multiplying the absolute weight value $|W^k_{i,j}|$ with the L2-norm of its corresponding input activation $\|X^{k-1}_j\|_2$ across the task dataset. This provides a more comprehensive measure of parameter importance than methods using only weight magnitude or gradients.
Next, we develop a model-agnostic parameter allocation algorithm that distributes trainable parameters evenly throughout the network. Rather than selecting parameters globally (which often concentrates them in top layers), our approach allocates a fixed budget to each neuron by selecting its top-K most important connections. This balanced distribution enables fine-tuning across all levels of feature abstraction while maintaining high parameter efficiency. Our experiments show that TaskEdge achieves comparable or better performance than existing methods while updating less than 0.1\% of parameters, making it ideal for resource-constrained edge devices.

\section{Related Works}

\subsection{Parameter-Efficient Fine-Tuning}
\label{sec:related_work:peft}

As large pre-trained models grow in scale, Parameter-Efficient Fine-Tuning (PEFT) approaches have emerged to reduce computational demands while maintaining performance \cite{hanParameterEfficientFineTuningLarge2024,huAdaptiveCommunicationsCollaborative2024, huFullSceneDomainGeneralization2024, huAgentsCoDriverLargeLanguage2024, huAgentsCoMergeLargeLanguage2024}. PEFT strategies fall into three categories: additive, selective, and reparameterization approaches.
\textit{Additive Fine-Tuning} incorporates new trainable components into frozen pre-trained models. Examples include \textit{Adapters} \cite{pmlr-v97-houlsby19a, huCollaborativePerceptionConnected2024,huCPGuardMaliciousAgent2024, hu2025cpguardnewparadigmmalicious, tao2025gcpguardedcollaborativeperception, tao2025directcpdirectedcollaborativeperception}, which integrate specialized modules within transformer architectures, and \textit{Prompt Tuning}, which augments input sequences with learnable vectors. While effective, these methods introduce inference overhead.
\textit{Selective Fine-Tuning} modifies only specific subsets of existing model parameters using a binary mask \(\mathcal{M}\) to indicate which parameters will be updated. This approach avoids increasing the model's overall complexity.
\textit{Reparameterization} restructures model parameters into low-rank formulations for efficient training. LoRA \cite{hu2022lora} applies trainable low-rank matrices to transformer weights, enabling compact storage of task-specific adaptations. Recent work on SPT \cite{heSensitivityAwareVisualParameterEfficient2023} combines sparse tuning with LoRA, demonstrating how focusing on task-relevant weights can deliver significant performance gains. However, existing methodologies generally lack task-awareness in parameter selection.

\subsection{Large Pre-Trained Model Task Adaptation}

To effectively adapt pre-trained large language models (LLMs) to downstream tasks, researchers have explored diverse strategies targeting optimal pre-training data, model architectures, and parameter configurations \cite{lin2025leo, lin2024split, lin2023pushing, lin2024splitlora, fang2024ic3m}. For instance, Cui \textit{et al.} \cite{8578530} leverage Earth Mover's Distance to select task-specific pre-training data by identifying the top $K$ source domain classes most similar to the target task. Yoon \textit{et al.} \cite{pmlr-v119-yoon20a} introduce a reinforcement learning framework to dynamically weight classes in the source domain. Other approaches assess model transferability by analyzing inter-class covariance between source and target data or measuring conditional cross-entropy across domains \cite{9009545, pmlr-v119-nguyen20b, tao2025gcpguardedcollaborativeperception, fangPACPPriorityAwareCollaborative2024}. Recent advances focus on identifying optimal subsets of pre-trained weights to fine-tune while freezing others. Zhang \textit{et al.} \cite{zhangGradientbasedParameterSelection2024} propose gradient-based parameter selection (GPS), which prioritizes tunable parameters via gradient analysis. Similarly, Fu \textit{et al.} \cite{fuEffectivenessParameterEfficientFineTuning2023} develop a second-order approximation method (SAM) to estimate the Hessian matrix in the loss function's Taylor expansion, enabling more informed parameter selection. In contrast to these methods, our approach adaptively selects task-aware tunable parameters by jointly analyzing weight and activation patterns. This mechanism offers a computationally efficient solution while maintaining simplicity, distinguishing it from prior gradient- or approximation-based techniques.

\subsection{Edge Computing for Large Pre-trained Models}

The demand for edge-deployed LLMs is driven by needs for faster response times, low latency, offline functionality, privacy, and personalized experiences, as seen in applications on smartphones, wearables, and smart home assistants \cite{10835069, lin2024efficient, lin2024fedsn, lin2024adaptsfl, lin2024hierarchical, qu2024trimcachingparametersharingedgecaching, 10630945}. The background work highlights the widening gap between LLM complexity and edge device capabilities, with much research focusing on efficient LLMs and edge computing optimizations including quantization \cite{10.5555/3600270.3602468}, pruning \cite{10.5555/3618408.3618822}, knowledge distillation \cite{10.5555/3495724.3496209}, and low-rank approximations. For example, Dettmers \textit{et al.} \cite{10.5555/3600270.3602468} introduced a novel Int8 quantization method for large language models, combining vector-wise quantization and mixed-precision decomposition to reduce memory usage by half while maintaining full precision performance, enabling inference of models up to 175 billion parameters on consumer GPUs without degradation. Franter \textit{et al.} \cite{10.5555/3618408.3618822} proposed SparseGPT, a one-shot pruning method that efficiently compresses massive GPT-family language models like OPT-175B and BLOOM-176B to 50-60\% sparsity without retraining, achieving minimal accuracy loss by solving large-scale sparse regression problems in under 4.5 hours on a single GPU. In addition, Wang \textit{et al.} \cite{10.5555/3495724.3496209} proposed a task-agnostic distillation method that compresses large Transformer-based language models like BERT into smaller models by deeply mimicking the self-attention module of the teacher's last layer, introducing value-relation transfer, and optionally using a teacher assistant, achieving over 99\% accuracy retention on SQuAD 2.0 and GLUE tasks with 50\% fewer parameters and computations.

\section{Method}

\subsection{Preliminaries}

In this section, we briefly introduce the sparse fine-tuning method. Given a downstream task dataset $\mathcal{D} = \{(x_i, y_i)\}_{i=1}^N$, the objective of sparse fine-tuning is to adapt the pre-trained model to the downstream task with the sparsity constraint on the tunable weights $\mathcal{W}\in \mathbb{R}^{d_1\times d_2}$. The objective function of sparse fine-tuning is defined as:
\begin{equation}
    \begin{aligned}
    &\min_{\mathcal{W}'} \frac{1}{N}\sum_{i=1}^N \mathcal{L}\left(f(\theta(x_i); \mathcal{W}'), y_i\right), \\ 
    &\ \text{s.t.} \quad \mathcal{W}' = \mathcal{W} \odot \mathcal{M},
    \end{aligned}\label{eq:sparse_tuning}
\end{equation}
where $\mathcal{L}$ is the loss function of the downstream task, $\odot$ is the element-wise multiplication, and $\mathcal{M}\in\{0, 1\}^{d_1\times d_2}$ is the binary mask matrix with the same shape as $\mathcal{W}$. 
The binary mask matrix $\mathcal{M}$ can be determined in multiple ways: it can be set as a fixed hyperparameter, calculated in advance using heuristic methods like pre-pruning, or learned adaptively during the end-to-end fine-tuning process.

\subsection{Task-Aware Weights Analysis}
In sparse fine-tuning, a critical challenge lies in determining the optimal mask matrix $\mathcal{M}$ that satisfies the sparsity constraint while maximizing model performance. The selection of this mask matrix directly impacts both the model's effectiveness on downstream tasks and its training efficiency. The key is to identify which parameters are most relevant for the target task while maintaining the desired level of sparsity. Previous methods use different criteria to select the important parameters, such as the gradient of the loss function with respect to the weights \cite{zhangGradientbasedParameterSelection2024,fuEffectivenessParameterEfficientFineTuning2023}, and the pre-defined heuristic rules.
However, these methods have their own limitations. For example, the gradient-based methods require computing the gradient with respect to all the weights in the model, which can be computationally expensive. The pre-defined heuristic rules may not be optimal for all tasks and models. 

To address this gap, we propose a principled approach derived from statistical learning theory. Let us formalize this by considering a linear regression model with input features $\mathbf{x} \in \mathbb{R}^d$ and weight vector $\mathbf{w} \in \mathbb{R}^d$. The model's prediction is given by $\hat{y} = \mathbf{w}^T\mathbf{x}$. When adapting this model to a new task with dataset $\mathcal{D}' = \{(\mathbf{x}_i', y_i')\}_{i=1}^{N'}$, the objective is to optimize $\mathbf{w}$ such that the model minimizes the mean squared error on the target task. For computationally efficient adaptation, we aim to identify and update only a subset of weights $\mathcal{S} \subset \{1,2,\ldots,d\}$ that maximizes task performance while satisfying $|\mathcal{S}| \leq k$ for some budget constraint $k$. The contribution of the $i$-th weight to the model's output can be expressed as $w_i x_i$, which depends on two factors: (1) the magnitude $|w_i|$ and (2) the distribution of the corresponding feature value $x_i$ across the task dataset. Mathematically, if $|w_i|$ is large but $\|x_i\|_2 = \sqrt{\sum_{j=1}^{N'} (x_{ji}')^2}$ is consistently small, the product $w_i x_i$ will have minimal impact on the prediction. Conversely, if $|w_i|$ is small but $\|x_i\|_2$ is large, the contribution will also be limited. Therefore, the optimal parameter selection should consider both the weight magnitudes and the statistical properties of the corresponding input features across the specific task dataset.

Based on this insight, we propose a task-aware importance score calculation method for weights in neural networks. For the \(k\)-th layer with weight matrix 
\(\mathcal{W}^k \in \mathbb{R}^{d_{\text{out}} \times d_{\text{in}}}\) 
and hidden state matrix \(\mathcal{X}^{k-1} \in \mathbb{R}^{T \times d_{\text{in}}}\) from the previous layer
(where \(T\) represents the number of tokens in the downstream task dataset \(\mathcal{X}\)),
we compute an importance score matrix 
\(\mathcal{S}^k \in \mathbb{R}^{d_{\text{out}} \times d_{\text{in}}}\).
For any weight element at position \((i,j)\), its importance score \(\mathcal{S}_{i,j}\) is defined as:
\begin{equation}
    \mathcal{S}_{i,j} = \left| \mathcal{W}^k_{i,j} \right| \cdot \left\| \mathcal{X}^{k-1}_j \right\|_2.
\end{equation}
Here, \(\left| \cdot \right|\) denotes the absolute value, and \(\left\| \mathcal{X}^{k-1}_j \right\|_2\) computes the \(l_2\)-norm of the \(j\)-th input feature across all tokens. This formulation captures both the weight magnitude and the statistical significance of the corresponding input feature, providing a more comprehensive measure of parameter importance for the specific downstream task.

\begin{algorithm}[t]
    \caption{TaskEdge}
    \begin{algorithmic}[1]
    \State \textbf{Input:} Pre-trained model weights $\mathcal{W}$, dataset $\mathcal{D}$, sparsity constraint $\lambda$
    \State \textbf{Output:} Fine-tuned weight matrix $\mathcal{W}'$
    
    \State \text{// Step 1: Forward pass to collect activation statistics}
    \For{each batch $(\mathbf{x}, \mathbf{y}) \in \mathcal{D}$}
        \State Perform forward pass through the model
        \State Record hidden state activations $\mathcal{X}^{(k)}$ for each layer
    \EndFor
    
    \State \text{// Step 2: Compute task-aware importance scores}
    \For{each layer $k$ with weight matrix $\mathcal{W}^{(k)}$}
        \State Compute $\|\mathcal{X}^{(k)}_j\|_2$ for each input feature $j$
        \For{each weight element $(i,j)$ in $\mathcal{W}^{(k)}$}
            \State $\mathcal{S}_{i,j} = |\mathcal{W}^{(k)}_{i,j}| \cdot \|\mathcal{X}^{(k)}_j\|_2$
        \EndFor
    \EndFor
    
    \State \text{// Step 3: Allocate trainable weights}
    \For{each neuron $n$ in the network}
        \State Select top-$K$ weights with highest $\mathcal{S}_{i,j}$ 
        \State Create binary mask $\mathcal{M}$ based on selected weights
    \EndFor
    
    \State \text{// Step 4: Fine-tune with sparse updates}
    \For{each epoch}
        \State Fine-tune using $\mathcal{W}' = \mathcal{W} - \gamma \nabla \mathcal{W} \odot \mathcal{M}$
    \EndFor
    \State \Return $\mathcal{W}'$
    \end{algorithmic}
\end{algorithm}

\subsection{Model-Agnostic Trainable Weights Allocation}

After computing the task-specific importance score, we need to allocate the trainable weights. A straightforward approach would be selecting a certain percentage of parameters with the highest importance scores from the entire model. However, this method has limitations since the selected parameters tend to concentrate in the top layers. Fine-tuning only these top-layer parameters proves insufficient, especially when there are significant differences between upstream and downstream task distributions that require adjusting more granular features in shallower layers.

On the contrary, when fine-tuning a pre-trained model, the adjustable parameters should be distributed throughout the network rather than concentrated in specific layers. This approach enables the model to adapt information stored at different levels of granularity to better fit downstream tasks. To achieve this, we propose a \textit{model-agnostic trainable weights allocation method}. Specifically, for each neuron in the network, we select the top-K weights with the highest importance scores. 

\textit{Discussion.} This neuron-level selection strategy ensures that every neuron can fine-tune its activation state rather than only adjusting high-level information in the top layers. Our approach enables fine-tuning of detailed information stored in each neuron, leading to better adaptation to downstream tasks. The process involves calculating importance scores $\mathcal{S}$ of weights $\mathcal{W}$ for a specific task, then selecting the top-K connections with the highest importance scores among all input connections for each neuron. This selection procedure not only identifies task-critical parameters but also allows comprehensive model adaptation across all layers. An additional advantage is its model-agnostic nature, making it readily applicable to various model architectures like Transformers and CNNs without requiring model-specific modifications.

\textit{Integration with Structured Sparsity.} Our approach, which has primarily focused on unstructured sparsity patterns, can be readily adapted to support structured N:M sparsity formats. In structured N:M sparsity, for any sequence of M adjacent weights, a maximum of N weights are allowed to be non-zero. This format is particularly beneficial as it enables acceleration using NVIDIA's specialized sparse tensor cores during matrix multiplication operations. To support N:M structured sparsity in our framework, we simply modify our weight selection process to apply our importance metric across groups of M consecutive weights at a time when determining which weights to keep active for each output neuron.

\begin{table*}[t]
    \centering
    \resizebox{\textwidth}{!}{
        \begin{tabular}{c|ccccccc|cccc|cccccccc|c}
        \toprule
         \multirow{2}{*} & \multicolumn{7}{c|}{Natural}                                       & \multicolumn{4}{c|}{Specialized}                   & \multicolumn{8}{c|}{Structured}                                                                                      &     \\
        \midrule
         & \rotatebox{90}{CIFAR-100} & \rotatebox{90}{Caltech101} & \rotatebox{90}{DTD} & \rotatebox{90}{Flowers102} & \rotatebox{90}{Pets} & \rotatebox{90}{SVHN} & \rotatebox{90}{Sun397} & \rotatebox{90}{Patch Camelyon} & \rotatebox{90}{EuroSAT} & \rotatebox{90}{Resisc45} & \rotatebox{90}{Retinopathy} & \rotatebox{90}{Clevr/count} & \rotatebox{90}{Clevr/distance} & \rotatebox{90}{DMLab} & \rotatebox{90}{KITTI/distance} & \rotatebox{90}{dSprites/loc} & \rotatebox{90}{dSprites/ori} & \rotatebox{90}{SmallNORB/azi} & \rotatebox{90}{SmallNORB/ele} & \rotatebox{90}{Mean Params. (\%)} \\
        \midrule
        Full~\cite{10.1007/978-3-031-19827-4_41}                            & 68.9      & 87.7       & 64.3 & 97.2       & 86.9 & \textbf{87.4} & 38.8   & 79.7           & 95.7    & 84.2     & 73.9        & 56.3        & 58.6           & 41.7  & 65.5           & 57.5         & 46.7         & 25.7          & 29.1              & 100.00           \\
        Linear~\cite{10.1007/978-3-031-19827-4_41}                          & 63.4      & 85.0       & 64.3 & 97.0       & 86.3 & 36.6 & 51.0   & 78.5           & 87.5    & 68.6     & 74.0        & 34.3        & 30.6           & 33.2  & 55.4           & 12.5         & 20.0         & 9.6           & 19.2               & 0.05             \\
        Bias~\cite{ben-zaken-etal-2022-bitfit}                            & 72.8      & 87.0       & 59.2 & 97.5       & 85.3 & 59.9 & 51.4   & 78.7           & 91.6    & 72.9     & 69.8        & 61.5        & 55.6           & 32.4  & 55.9           & 66.6         & 40.0         & 15.7          & 25.1               & 0.16             \\
        Adapter~\cite{pmlr-v97-houlsby19a}                         & 74.1      & 86.1       & 63.2 & 97.7       & 87.0 & 34.6 & 50.8   & 76.3           & 88.0    & 73.1     & 70.5        & 45.7        & 37.4           & 31.2  & 53.2           & 30.3         & 25.4         & 13.8          & 22.1              & 0.31             \\
        LoRA~\cite{huLORALOWRANKADAPTATION2022}                         & 68.1      & 91.4       & 69.8 & 99.0       & 90.5 & 86.4 & \textbf{53.1}   & \textbf{85.1}          & 95.8    & 84.7     & 74.2        & \textbf{83.0}        & \textbf{66.9}           & \textbf{50.4}  & \textbf{81.4}           & \textbf{80.2}& 46.6         & 32.2          & 41.1             & 0.90             \\
        LoRA~\cite{huLORALOWRANKADAPTATION2022}                         & 68.1      & 91.4       & 69.8 & 99.0       & 90.5 & 86.4 & \textbf{53.1}   & \textbf{85.1}          & 95.8    & 84.7     & 74.2        & \textbf{83.0}        & \textbf{66.9}           & \textbf{50.4}  & \textbf{81.4}           & \textbf{80.2}& 46.6         & 32.2          & 41.1             & 0.90             \\
        VPT-Shallow~\cite{10.1007/978-3-031-19827-4_41}                     & 77.7      & 86.9       & 62.6 & 97.5       & 87.3 & 74.5 & 51.2   & 78.2           & 92.0    & 75.6     & 72.9        & 50.5        & 58.6           & 40.5  & 67.1           & 68.7         & 36.1         & 20.2          & 34.1               & 0.13             \\
        VPT-Deep~\cite{10.1007/978-3-031-19827-4_41}                        & \textbf{78.8}      & 90.8       & 65.8 & 98.0       & 88.3 & 78.1 & 49.6   & 81.8           & \textbf{96.1}    & 83.4     & 68.4        & 68.5        & 60.0           & 46.5  & 72.8           & 73.6         & \textbf{47.9}         & \textbf{32.9}          & \textbf{37.8}               & 0.70             \\
        \midrule
        
        TaskEdge (Ours)  & 61.1 & \textbf{91.6}&\textbf{74.3} &\textbf{99.1} & \textbf{91.5} & 82.6 &  50.4& 81.9 & 94.4 & \textbf{85.6} & \textbf{74.8} &45.8 & 46.3 & 39.35 & 68.8 & 52.00 &37.9  & 19.7 & 27.0 &  0.09  \\
        \bottomrule
        \end{tabular}
    }
    \caption{Comparision results on VTAB-1k with pre-trained ViT-B/16 models on ImageNet-21K dataset.}
    \label{tab:vtab}
    \vspace{-5mm}
\end{table*}

\subsection{Combining with LoRA}

Our method can be seamlessly applied to the reparameterization methods stated in Section \ref{sec:related_work:peft}. Here we take LoRA as an example to illustrate its plug-and-play superiority.

Given a pre-trained weight matrix $\mathcal{W}_0\in \mathbb{R}^{d_1\times d_2}$, the LoRA introduces two trainable rank-deficient matrix $\mathcal{B}\in \mathbb{R}^{d_1\times r}$ and $\mathcal{A}\in \mathbb{R}^{r\times d_2}$, where $r$ is the rank parameter, $r\ll \min(d_1, d_2)$. The weight matrix $\mathcal{W}_0$ can be reparameterized as:
\begin{equation}
    \mathcal{W} = \mathcal{W}_0 + \Delta \mathcal{W} = \mathcal{W}_0 + \mathcal{B}\times\mathcal{A},
\end{equation}
where $\times$ is the dot product of two matrices. In this way, the number of trainable parameters is reduced from $d_1\cdot d_2$ to $(d_1+d_2)\cdot r$, which is a significant reduction when $r$ is a small constant. Therefore, LoRA can achieve high training efficiency and low memory usage.

To integrate our method with LoRA, the most straightforward way is to introduce two binary mask matrices $\mathcal{M}_B\in\{0, 1\}^{d_1\times r}$ and $\mathcal{M}_A\in\{0, 1\}^{r\times d_2}$ to the trainable parameters $\mathcal{B}$ and $\mathcal{A}$, respectively. Then, the tunable parameters $\Delta \mathcal{W}$ can be written as:
\begin{equation}
    \Delta \mathcal{W} = (\mathcal{M}_B\odot \mathcal{B})\times (\mathcal{M}_A\odot \mathcal{A}),
    \label{eq:lora_mask}
\end{equation}
where $\odot$ is the element-wise multiplication. Then, Eq. \ref{eq:lora_mask} can be rewritten as:
\begin{equation}
    \Delta \mathcal{W} =  (\mathcal{B}\times \mathcal{A}) \odot (\mathcal{M}_B\times \mathcal{M}_A).
    \label{eq:lora_mask_final}
\end{equation}
We can treat $\mathcal{M}_B\times \mathcal{M}_A$ as a whole, which is consistent with the form in Eq. \ref{eq:sparse_tuning}. Therefore, we only need to introduce one mask matrix $\mathcal{M}\in\{0, 1\}^{d_1\times d_2}$ to achieve sparsification, which has the same dimension as the model's weight matrix $\mathcal{W}_0$. The LoRA with sparse fine-tuning can be formulated as:
\begin{equation}
    \mathcal{W} = \mathcal{W}_0 + \Delta \mathcal{W} = \mathcal{W}_0 + (\mathcal{B}\times \mathcal{A}) \odot \mathcal{M}.
    \label{eq:lora_sparse_tuning}
\end{equation}

\section{Experiments}

\subsection{Dataset and Baselines}

\textit{Dataset.} We leverage the Visual Task Adaptation Benchmark (VTAB-1k) \cite{zhai2020largescalestudyrepresentationlearning} as our dataset. It contains 19 sub-datasets for different vision tasks adaptation which are categorized into 3 groups: Natural, Specialized, and Structured. 

\textit{Baselines}. In order to verify the effectiveness of our proposed method, we compare it with several baselines, including Full \cite{10.1007/978-3-031-19827-4_41}, Linear \cite{10.1007/978-3-031-19827-4_41}, Bias \cite{ben-zaken-etal-2022-bitfit}, Adapter \cite{pmlr-v97-houlsby19a}, LoRA \cite{hu2022lora}, VPT \cite{10.1007/978-3-031-19827-4_41}, and Magnitude \cite{pmlr-v97-houlsby19a}. Full fine-tuning is the most commonly used protocol, which updates all parameters of the entire model during tuning. The Linear baseline involves updating only the linear layers of the model. The Bias method is a sparse-finetuning approach where only the bias terms of the model, or a subset of them, are modified. Adapter and VPT methods add new trainable parameters to the backbone of the model. LoRA is a reparameterization technique where weight updates are applied as low-rank matrices to the original weights.

\subsection{Implementation Details}
We employ ViT-B/16 \cite{dosovitskiy2021an}, which has been pre-trained on ImageNet-21k \cite{5206848}, as the backbone for our experiments.
Furthermore, we utilize the Adam optimizer with a cosine learning rate decay schedule to fine-tune the models over 100 epochs, incorporating a linear warm-up phase during the initial 10 epochs. All experiments are performed on 4 NVIDIA A5000 GPUs.

\begin{figure}[t]
    \centering
    \includegraphics[width=1\linewidth]{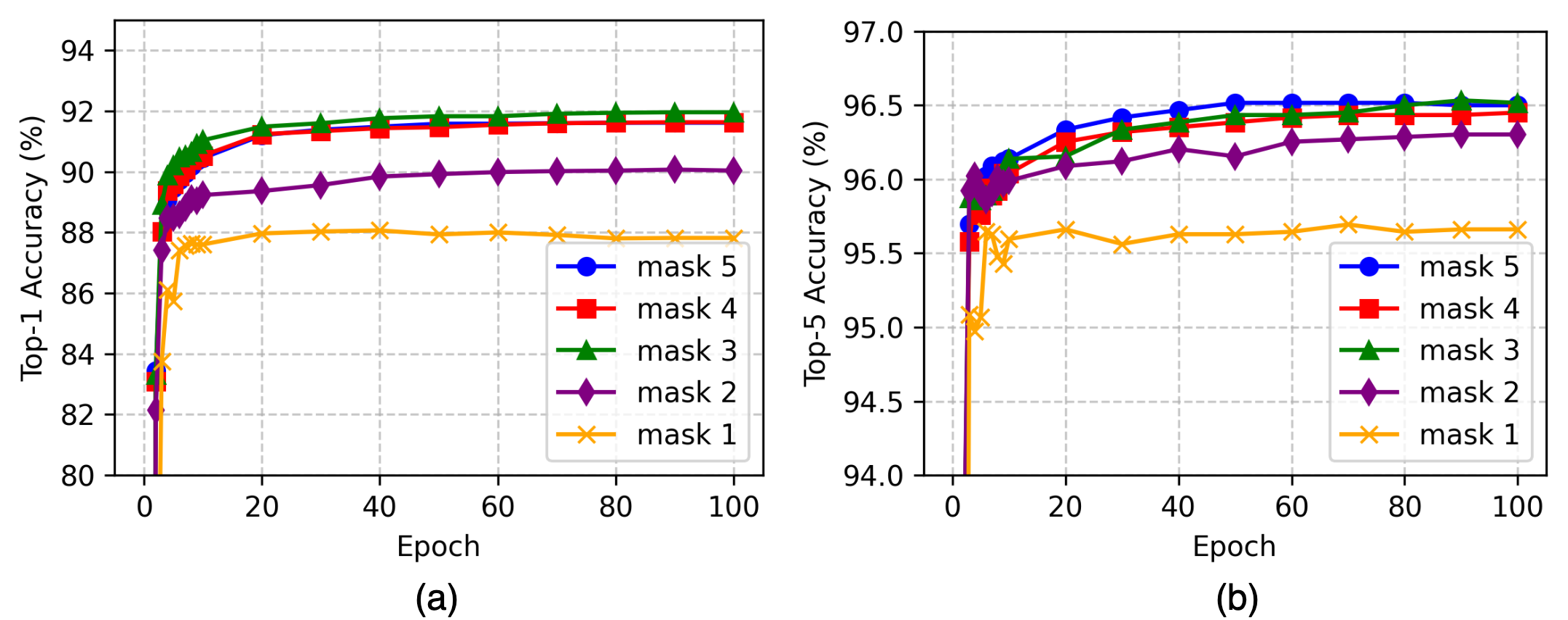}
    
    \caption{Epochs vs Accuracy}
    \label{fig:ablation_study1}
    \vspace{-5mm}
\end{figure}

\subsection{Comparison Results}

In this section, we present the comprehensive comparison results of our proposed method with the baselines. The results are shown in Tab. \ref{tab:vtab}. Our method achieves comparable or superior results with just 0.09\% of the parameters of the full model, demonstrating remarkable parameter efficiency. Specifically, TaskEdge outperforms all baselines on several natural datasets, including Caltech101, DTD, Flowers102, and Pets, achieving the highest accuracy of 91.6\%, 74.3\%, 99.1\%, and 91.5\% respectively. For specialized datasets, our method also demonstrates superior performance on Resisc45 and Retinopathy with accuracies of 85.6\% and 74.8\%.
When compared to LoRA, which uses 0.90\% of the parameters, our method achieves better results on 5 datasets while using only one-tenth of the trainable parameters. Similarly, when compared to VPT-Deep, which uses 0.70\% of the parameters, TaskEdge shows competitive performance across multiple datasets with significantly fewer parameters. These results highlight the effectiveness of our task-aware parameter allocation strategy in identifying the most critical parameters for specific downstream tasks.
Furthermore, we observe that TaskEdge maintains consistent performance across the diverse range of tasks in VTAB-1k, indicating its robustness and adaptability to different visual domains. The structured task category, which typically requires more complex reasoning capabilities, also benefits from our approach, with TaskEdge achieving competitive results while maintaining minimal parameter usage. This demonstrates that our method effectively captures task-specific information even for challenging visual reasoning tasks.

\subsection{Ablation Studies}

\begin{figure}[t]
    \centering
    \includegraphics[width=1\linewidth]{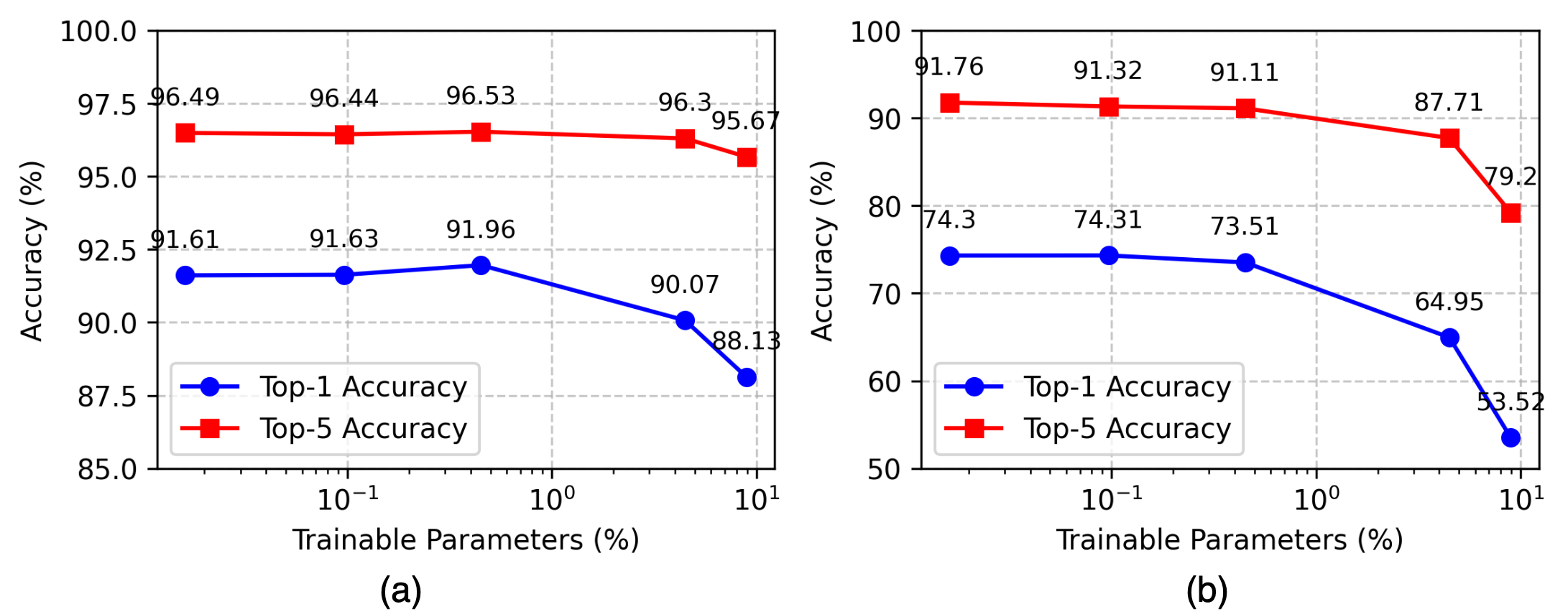}
    \caption{Trainable Parameters vs Accuracy}
    \label{fig:ablation_study2}
    \vspace{-5mm}
\end{figure}

In order to further analyze the performance of our method, we plot the accuracy vs epochs and trainable parameters vs accuracy in Fig. \ref{fig:ablation_study1} for the Caltech-101 dataset. We set the mask ratio to 91.06\%, 95.52\%, 99.55\%, 99.90\%, and 99.98\%, which are mask 1, 2, 3, 4, and 5 respectively. Here the masked parameters are the parameters that are not updated during the fine-tuning process. Fig. \ref{fig:ablation_study1}(a) shows the Top-1 accuracy and Fig. \ref{fig:ablation_study1}(b) shows the Top-5 accuracy. In Fig. \ref{fig:ablation_study2}(a) and Fig. \ref{fig:ablation_study2}(b), we plot the trainable parameters vs accuracy in Caltech-101 and DTD datasets, respectively.
The results show that the model converges around 20 epochs, and when the mask ratio is around 99\%, the model achieves the highest accuracy. When the trainable parameters increase, the accuracy decreases, because the training set is not large enough and the model is easy to overfit. The results of full fine-tuning in Tab \ref{tab:vtab} also show this phenomenon. However, our method can handle this problem to some extent by selecting the most important parameters to update and maintaining the model performance.

\section{Conclusion}
In this paper, we presented TaskEdge, a novel task-aware parameter-efficient fine-tuning framework designed specifically for edge computing environments. Our approach addresses the critical challenge of adapting large pre-trained models to specific downstream tasks while operating within the computational and memory constraints of edge devices. By intelligently allocating only the most effective parameters to target tasks, TaskEdge achieves remarkable efficiency without sacrificing performance.
Our experimental results on the Visual Task Adaptation Benchmark (VTAB-1k) demonstrate that TaskEdge achieves comparable or superior performance to existing parameter-efficient fine-tuning methods while using significantly fewer parameters (less than 0.1\% of the total parameters).

\section{Acknowledgment}
The research work described in this paper was conducted in the JC STEM Lab of Smart City funded by The Hong Kong Jockey Club Charities Trust under Contract 2023-0108.  The work was supported in part by the Hong Kong SAR Government under the Global STEM Professorship and Research Talent Hub. The work of S. Hu was  supported in part by the Hong Kong Innovation and Technology Commission under InnoHK Project CIMDA. The work of Y. Deng was supported in part by the National Natural Science Foundation of China under Grant No. 62301300.

\footnotesize
{\bibliography{ref, ref2}

\bibliographystyle{IEEEtran}
}

\end{document}